# Functional Localization Enforced Deep Anomaly Detection Using Fundus Images

Jan Benedikt Ruhland[1], Thorsten Papenbrock[2], Jan-Peter Sowa[3], Ali Canbay[3], Nicole Eter[4], Bernd Freisleben[2], Dominik Heider[5*]

1: Heinrich Heine University Düsseldorf, Faculty of Mathematics and Natural Sciences, Düsseldorf, Germany
2: Philipps University of Marburg, Department of Mathematics and Computer Science, Marburg, Germany
3: University Hospital Knappschaftskrankenhaus Bochum, Department of Internal Medicine, Germany
4: University of Münster, Department of Ophthalmology, Münster, Germany
5: University of Münster, Institute of Medical Informatics, Münster, Germany

*corresponding author: dominik.heider@uni-muenster.de

## Abstract

Reliable detection of retinal diseases from fundus images is challenged by the variability in imaging quality, subtle early-stage manifestations, and domain shift across datasets. In this study, we systematically evaluated a Vision Transformer (ViT) classifier under multiple augmentation and enhancement strategies across several heterogeneous public datasets, as well as the AEyeDB dataset, a high-quality fundus dataset created in-house and made available for the research community. The ViT demonstrated consistently strong performance, with accuracies ranging from 0.789 to 0.843 across datasets and diseases. Diabetic retinopathy and age-related macular degeneration were detected reliably, whereas glaucoma remained the most frequently misclassified disease. Geometric and color augmentations provided the most stable improvements, while histogram equalization benefited datasets dominated by structural subtlety. Laplacian enhancement reduced performance across different settings.

On the Papila dataset, the ViT with geometric augmentation achieved an AUC of 0.91, outperforming previously reported convolutional ensemble baselines (AUC of 0.87), underscoring the advantages of transformer architectures and multi-dataset training. To complement the classifier, we developed a GANomaly-based anomaly detector, achieving an AUC of 0.76 while providing inherent reconstruction-based explainability and robust generalization to unseen data. Probabilistic calibration using GUESS enabled threshold-independent decision support for future clinical implementation.

Dataset-level analysis highlighted that acquisition consistency substantially impacts model robustness. Curated datasets with manual quality control yielded the highest performance. Overall, our results support the combined use of transformer-based classifiers and principled anomaly detection for fundus analysis and motivate future work on domain adaptation and improved calibration through larger healthy cohorts.

# 1. Introduction

Fundus imaging is a non-invasive, cost-effective, and widely accessible modality used for visualizing the retina and its surrounding structures. It plays an essential role in the early detection and monitoring of retinal disorders such as age-related macular degeneration (AMD) [1, 2], diabetic retinopathy (DR) [3], and glaucoma [4]. These diseases exhibit distinct visual patterns, ranging from microaneurysms and hemorrhages to optic disc cupping and macular degeneration, which can be captured in fundus photographs. However, the visibility and distinctiveness of these features vary strongly. While some abnormalities are easily recognizable, even by non-specialists, others manifest as subtle structural or textural deviations that may elude expert ophthalmologists. Nevertheless, the affordability, non-invasive nature, and scalability of fundus imaging makes it indispensable for both clinical diagnosis and large-scale screening programs.

Beyond its clinical significance, fundus imaging has become a central focus in medical image analysis research, driving progress in vessel segmentation, lesion localization, synthetic data generation, and automated disease classification [5, 6, 7, 8]. Recent studies have demonstrated that retinal fundus images can be used not only to diagnose specific ophthalmic diseases but also to infer systemic health conditions such as cardiovascular risk, anemia, and even biological aging [9, 10, 11]. Moreover, datasets such as JSIEC [6] reveal that fundus images can display features corresponding to more than 39 distinct pathological abnormalities, each representing a unique manifestation of retinal or systemic disease. This vast pathological diversity underscores a critical challenge for conventional supervised classification approaches. Training deep neural networks to accurately classify dozens of disease types requires extensive labeled datasets and suffers from severe class imbalance. Furthermore, rare or previously unseen abnormalities may be misclassified, limiting the reliability and generalizability of such systems in real-world clinical settings.

Deep learning models designed for fundus image analysis, including convolutional neural networks (CNNs) and transformer-based architectures, have achieved remarkable accuracy in recent studies [12, 13, 14]. However, their deployment in clinical environments is constrained by several factors. First, these models are computationally demanding, requiring high-end GPUs and large memory footprints, and second, they often depend on cloud-based processing, which raises concerns about patient privacy and data security. Additionally, despite their predictive power, such models generally act as black boxes. They output categorical labels or probabilities without explaining the underlying rationale behind each decision. This lack of interpretability hinders clinical validation and erodes trust among medical practitioners, who must understand and verify the reasoning behind AI-driven assessments before integrating them into patient care.

Recent advances in explainable artificial intelligence (AI) aim to address these challenges by identifying and visualizing the image regions most influential in model predictions. Frameworks such as SHAP [15] and DeepLIFT [16] quantify feature importance, while attention-based and gradient-based visualization methods generate saliency maps or contribution maps highlighting diagnostically relevant areas [17, 18]. By localizing key retinal structures or lesions that contribute to the final decision, these methods enhance transparency, facilitate clinical interpretability, and allow for more informed validation of AI-generated results.

Given the large number of potential retinal pathologies and the limitations of fixed-category classification, anomaly detection offers a more scalable and clinically realistic alternative. Instead of explicitly classifying every possible disease, anomaly detection frameworks learn the normal retinal appearance and identify deviations indicative of a pathology. This paradigm is particularly well-suited for medical imaging, where the spectrum of disease manifestations is vast and often unpredictable, e.g., for rare diseases.

In this study, we present a comprehensive framework for explainable retinal image analysis that integrates transformer-based classification, reconstruction-driven anomaly detection, and probabilistic calibration. Our work systematically evaluates the performance and robustness of a Vision Transformer classifier across multiple augmentation strategies, investigates a constrained attention mechanism for embedded localization, and develops an anomaly detection pipeline based on GANomaly [19] combined with GUESS [20] calibration. In addition, we introduce AEyeDB, a newly curated, high-quality fundus image dataset created by our group and made available to the research community upon request.

The main contributions of this research are:

1. Systematic Evaluation of Transformer-Based Fundus Classification

   We provide a detailed assessment of ViT performance under geometric, chromatic, and contrast-enhancing augmentation strategies across multiple heterogeneous datasets. Our results show that while geometric and color augmentations offer stable improvements, histogram equalization and Laplacian enhancement can degrade performance, providing practical guidance for preprocessing choices in fundus classification pipelines.

2. Embedded Attention Mechanism for Inherent Localization

   We incorporate a constrained attention-based masking strategy directly into the ViT architecture to promote inherent spatial interpretability. Although this mechanism did not directly improve classification accuracy, it produced consistent and physiologically meaningful attention maps, demonstrating the feasibility of integrating functional localization into end-to-end training without relying on post-hoc saliency methods.

3. Reconstruction-Based Anomaly Detection with Explainability

   We extend the GANomaly framework to provide interpretable anomaly maps derived from reconstruction error distributions. Although additional loss terms (mask loss, KL regularization) did not improve performance, the baseline model achieved an AUC of 0.76 and generalized well to unseen data. The reconstruction maps reliably highlighted pathological structures, offering a principled and transparent alternative to black-box classifier outputs.

4. Probabilistic Calibration With GUESS

   We employ the GUESS framework to transform raw anomaly scores into calibrated abnormality probabilities. This threshold-independent approach produces well-ranked probabilistic outputs and facilitates clinically meaningful decision support. The use of AUC as the primary metric aligns with the goal of evaluating probability ordering rather than thresholded classification.

5. Introduction of the AEyeDB Dataset

We release AEyeDB, a curated, high-quality fundus imaging dataset collected and annotated by our team. The dataset will be accessible upon request to researchers who provide a brief description of their intended research aims, promoting transparency, reproducibility, and community collaboration.

## Research in Context

The development of multi-head attention in large language models [21] has inspired the adaptation of attention mechanisms to the image domain through architectures such as the Vision Transformer [22]. ViTs leverage self-attention to assign varying importance to different spatial regions of an image, allowing the model to dynamically focus on informative features while ignoring irrelevant areas. This inherent ability to localize salient regions makes attention mechanisms particularly attractive for medical imaging, where understanding why a model predicts a certain outcome is as important as the prediction itself.

Attention mechanisms have been widely applied in medical imaging tasks, including fundus disease classification [6, 23], chest X-ray interpretation [24], and skin lesion detection [25]. In these applications, attention maps provide interpretable visualizations that highlight regions contributing most to a model's decision, thereby improving clinician trust and enabling better validation of AI-driven diagnoses. In fundus imaging specifically, prior studies have focused primarily on disease classification and post-hoc interpretability. For instance, an ensemble of convolutional neural networks (CNNs) was used to classify 39 fundus pathologies, achieving a weighted F1 score of 92.3% while employing SHAP values to enhance model explainability [6]. Similarly, ViT-based architectures have been employed to generate attention maps for visual explanations in retinal disease classification [26]. These works typically rely on self-attention mechanisms as introduced by Vaswani et al. [21], where the focus of the network is determined dynamically but independently of the anomaly detection objective.

While these methods provide useful post-hoc interpretations, they often require additional processing after training, and their ability to identify subtle or previously unseen pathological features remains limited. Traditional interpretability techniques such as SHAP [15], CAM [27], and Grad-CAM [28] quantify feature contributions after model training, which can create a gap between the learning process and the explanatory outputs. In contrast, our approach integrates an explainable masking mechanism directly into the network architecture, making interpretability an inherent component of model learning. This design allows the model to learn which regions are diagnostically relevant during training, rather than inferring importance post hoc.

Moreover, while most prior studies have focused exclusively on either classification or localization, our framework combines both objectives in a unified architecture. We introduce sparsity through an attention mask that guides the learning process, enforcing the model to concentrate on the most informative regions while suppressing irrelevant features. This attention-guided mechanism is then applied within a discriminative anomaly detection framework, enabling the identification of pathological anomalies even in cases where explicit class labels are unavailable or diseases are rare.

By integrating functional localization with anomaly detection, our method addresses a key limitation of prior work: the difficulty to generalize across a large spectrum of abnormalities, which is critical given that biomarkers like fundus images can contain over 39 distinct pathological patterns.

Finally, our study investigates whether the incorporation of learned attention masks enhances both performance and interpretability of disease detection models. By embedding localization directly within the learning process and coupling it with anomaly detection, we provide a scalable, transparent, and clinically relevant framework for fundus image analysis, distinct from traditional post-hoc attention or classification-only approaches.

## 2. Materials and Methods

### 2.1 Proposed Models

The proposed framework builds upon the Vision Transformer architecture and extends it through a constrained attention mechanism for interpretable localization and anomaly detection. As a baseline model, we employ the ViT-B/16 configuration containing approximately 86 million trainable parameters [22]. The pretrained model is first used to evaluate domain shifts and the effect of various data augmentation strategies on fundus images. The most effective configuration is then adapted to train a segmentation network with constrained attention, allowing the model to highlight diagnostically relevant regions during inference.

For spatial localization, we employ a U-Net based segmentation network endowed with adaptive normalization [29] and self-attention [21] in the bottleneck region (described in Appendix A). The network outputs a single-channel mask $M=g_\phi(x)\in[0,1]$ through a sigmoid activation, where each pixel value represents the estimated importance of that region. This mask is applied elementwise to the input image, producing a masked image $x'=M\odot x$, which is subsequently passed into the ViT classifier $f_\theta$ that has been pretrained on multi-class fundus classification. The model prediction is expressed as $y'=f_\theta(x')$.

The segmentation network $g_\phi$ is optimized using a composite loss function that combines the standard cross-entropy loss from the classifier with an $L_1$ regularization term to encourage sparsity in the generated masks. The total loss is formulated as

$$L=L_{CE}(f_\theta(g_\phi(x)\odot x), y)+\lambda\|M\|_1 \quad (1)$$

where the cross-entropy loss $L_{CE}$ enforces prediction accuracy relative to the ground truth, and the regularization coefficient $\lambda$ controls the trade-off between classification fidelity and mask sparsity. By minimizing this loss, the network learns to identify minimal yet sufficient regions required for accurate classification, effectively removing irrelevant background and noise. This process serves as an embedded preprocessing step that enhances both interpretability and robustness.

The resulting attention masks can be interpreted as learned importance maps that highlight regions influencing the classifier's decision, similar in spirit to post-hoc interpretability methods such as SHAP and Grad-CAM [15, 28]. However, unlike these approaches, which operate after model training, our constrained attention mechanism is fully integrated into the training process, making localization an intrinsic part of learning.

Building on this foundation, the attention mechanism is extended to an anomaly detection framework that leverages the generative-adversarial architecture of GANomaly [19]. The model follows an encoder-decoder-encoder structure in which the input image is first encoded into a latent representation, reconstructed by the decoder, and then re-encoded to assess reconstruction consistency. Deviations between original and reconstructed latent features form the basis of the anomaly score A(x). In the classical formulation, GANomaly employs three loss components: an $L_2$ adversarial feature map loss, an $L_1$ reconstruction loss, and an $L_2$ latent consistency loss.

Inspired by recent developments in variational autoencoders [30] and diffusion models [31], we modify this architecture by introducing a Gaussian prior regularization on the latent space through a Kullback-Leibler (KL) divergence term. This additional constraint enforces a structured, stable latent distribution, improving reconstruction fidelity and anomaly discrimination [31]. Two configurations are therefore explored: the vanilla GANomaly model and the KL-regularized variant. The better configuration is then also trained in conjunction with the constrained attention mask introduced earlier, allowing us to evaluate how attention-guided learning influences anomaly detection performance.

The integration of the attention mask into the autoencoder training pipeline further enhances the reconstruction process by emphasizing diagnostically relevant retinal regions. This targeted focus reduces redundancy and prevents background features from diluting the anomaly signal. Consequently, the generative model learns to reconstruct only the most informative areas, improving the spatial alignment between detected anomalies and actual pathological regions in the fundus images.

Technically, the integration is achieved by introducing an additional reconstruction loss term applied selectively to the top 90th percentile of the attention-weighted image intensities. This selective penalization ensures that the model prioritizes reconstruction accuracy in regions deemed clinically significant, while disregarding low-importance background areas.

Finally, the computed anomaly score A(x) is calibrated using the GUESS framework [20]. Unlike conventional GANomaly methods that normalize scores empirically over healthy and abnormal datasets, GUESS provides a probabilistic calibration based on Bayes' theorem, yielding P(pathology|A(x)). This transforms raw anomaly scores into interpretable probabilities of pathology, thereby enhancing clinical interpretability and reliability.

Overall, this integrated approach combines discriminative attention, generative reconstruction, and probabilistic calibration in a unified pipeline. The ViT-based classifier establishes a robust and interpretable baseline, while the constrained attention mask enforces spatial focus. The KL-regularized GANomaly model captures deviations from the learned healthy distribution, and GUESS converts these deviations into probabilistic indicators of disease. Together, these elements form a cohesive system for interpretable, accurate, and generalizable anomaly detection in retinal fundus imaging.

## 2.2 Experimental Procedure

### 2.2.1. Fundus Image Datasets

We conducted a study at the Heinrich Heine University Düsseldorf involving 103 young adults (age≥18, written consent provided). Participants were informed on the potential risks of phototoxicity and epileptic triggering associated with the image recording process. Subsequently, they completed a questionnaire to collect data on age, gender, and preexisting conditions/diseases. Fundus images from the participants were captured using the Rodenstock FundusScope. The Beurer PO 80 was used to record the pulse waveform of the participants.

The average age in the study was 26.93 ± 7.18 years (see also Figure 1). Out of the 103 participants, 21 were female and 82 were male. In the provided questionnaire, 13 participants reported a pre-existing condition which might influence the retina recording, e.g., diabetes or hypertension. Five participants took part in the study multiple times, resulting in a total of 256 fundus images. For the analysis, we utilized a subset of 204 images to minimize potential biases arising from multiple images of the same individual and to exclude recordings of insufficient quality.

The study was approved by the ethical committee at Heinrich Heine University and registered (https://drks.de/search/de/trial/DRKS00033094) to ensure transparency and adherence to ethical research standards.

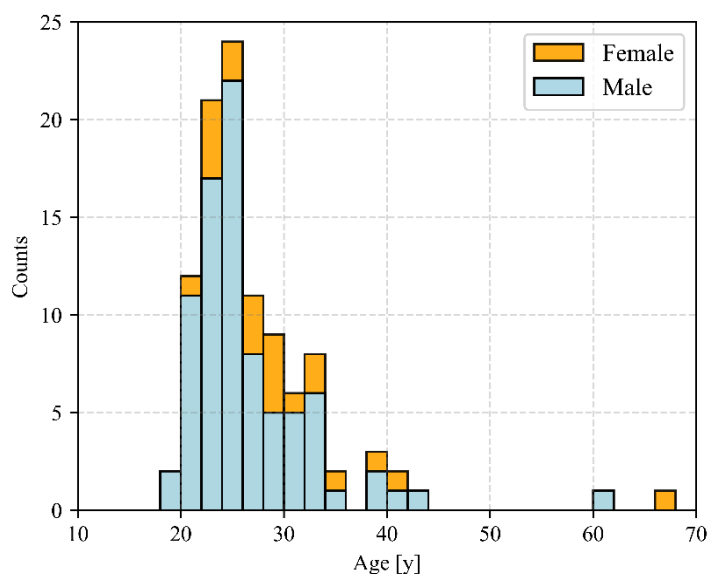

**Figure 1:** The figure illustrates the age distribution for the AEyeDB and the bin width is two years. Within each bin, the blue area indicates the proportion of male patients, while the orange area represents the fraction of female patients.

In addition to the AEyeDB dataset, several publicly available datasets were employed to enhance model generalizability and enable cross-dataset evaluation. These include FIVES [5], Mendeley [32], Papila [33], Messidor [34], and JSIEC [6, 35].

The FIVES dataset comprises 800 fundus images collected at the Ophthalmology Centre of the Second Affiliated Hospital of Zhejiang University. The images are evenly distributed across four diagnostic categories: healthy, diabetic retinopathy (DR), glaucoma, and age-related macular degeneration (AMD) (200 images per class). The dataset also provides image quality annotations and was recorded using a Topcon TRC-NW8 fundus camera under uniform imaging conditions. The consistent acquisition protocol makes FIVES a suitable external reference dataset for evaluating cross-domain performance.

The Mendeley dataset is a large compilation of publicly available retinal fundus images sourced from multiple clinical studies. It contains diverse pathological categories including diabetic retinopathy, glaucoma, macular scar, retinitis pigmentosa, and disc edema, with over 5,000 high-resolution images spanning healthy and diseased retinas. All images are labeled according to the predominant clinical diagnosis, providing a broad spectrum of retinal pathologies for model training and evaluation.

The Papila dataset focuses on the optic nerve head region and contains 420 stereo fundus image pairs (corresponding to 210 subjects). It includes both healthy controls and glaucoma patients, along with corresponding segmentation masks of the optic disc and cup regions.

The Messidor database consists of 1,200 fundus images, originally designed for benchmarking diabetic retinopathy detection algorithms. While the dataset itself lacks fine-grained lesion annotations, the Maples-DR extension [30] provides detailed expert-annotated masks for pathological features including blood vessels, drusen, exudates, hemorrhages, microaneurysms, and cotton wool spots.

To further evaluate the generalizability and robustness of the proposed model, we incorporated the JSIEC dataset as an independent test set. JSIEC contains over 209,000 fundus images representing a wide range of ocular conditions. However, only a limited subset is publicly available via Kaggle [34]. For this study, we selected images labeled as healthy, diabetic retinopathy (DR2), fibrosis, pathological myopia, optic atrophy, and laser spots to cover a representative spectrum of both common and rare retinal pathologies. Using JSIEC solely for testing ensures that the model's evaluation reflects realistic domain shifts and unseen disease types.

The distribution of fundus images across datasets and diagnostic categories after preprocessing is presented in table 1. This compilation integrates both in-house and public datasets, thereby enabling a comprehensive assessment of classification, localization, and anomaly detection performance.

**Table 1**: The table shows the distribution of fundus images given the underlying classification and the corresponding dataset after preprocessing.

| Classification | AEyeDB | FIVES | Mendeley | Papila | Messidor | JSIEC |
|---|---|---|---|---|---|---|
| Normal | 204 | 178 | 1024 | 333 | 480 | 38 |
| DR | - | 144 | 1509 | - | 119 | 49 |
| Glaucoma | - | 99 | 1349 | 87 | - | - |
| AMD | - | 171 | - | - | - | - |
| Macular Scar (MS) | - | - | 444 | - | - | - |
| Retinitis Pigmentosa (RP) | - | - | 139 | - | - | - |
| Disc Edema (DE) | - | - | 127 | - | - | - |
| Fibrosis | - | - | - | - | - | 10 |
| Pathological Myopia | - | - | - | - | - | 54 |
| Optic Atrophy | - | - | - | - | - | 12 |
| Laser Spots | - | - | - | - | - | 20 |

### 2.2.2. Fundus Image Preprocessing

The quality of fundus images can vary substantially due to environmental conditions, camera calibration, and subject-specific factors such as eye movement or pupil dilation. These variations often manifest as artifacts including poor illumination, blurring, color distortion, and low contrast, which can impair model performance and reliability. To ensure data consistency, we excluded all low-quality images from the FIVES dataset that were annotated as suboptimal according to the provided quality assessment criteria.

Fundus images also differ with respect to camera angle, field of view, and focal plane, all of which influence the visibility of key intraocular structures such as the optic nerve head, macula, and vascular arcades. Such variations can introduce domain shifts across datasets and affect diagnostic tasks, particularly in diseases like glaucoma, where the optic disc morphology is essential for accurate detection [32].

All images were normalized, center-cropped and resized to 224 × 224 pixels, aligning with the input resolution of the ViT-B/16 model. To systematically evaluate preprocessing and augmentation strategies for domain generalization, we implemented a stepwise procedure:

1. Baseline (No Augmentation): Images were normalized and directly fed into the network without additional transformations to establish a performance reference.

2. Geometric Augmentation: We applied random horizontal flipping, full rotations (random degree between 0° and 360°) and random translations up to 10% of image width and height to simulate variations in camera positioning and patient alignment.

3. Color and Blur Augmentation: Gaussian blur and color jittering were incorporated to simulate lighting inconsistencies and chromatic aberrations, improving generalization to real-world imaging conditions.

4. Contrast and Contour Enhancement: Finally, we evaluated the integration of histogram equalization [37] and Laplacian enhancement [38] within the strongest augmentation setup to emphasize vascular and structural details. These techniques were compared to assess their effect on model convergence and localization quality.

To ensure robust evaluation and prevent overfitting, we randomly reserved 15% of each dataset for validation and 15% for testing, stratified across all disease classes. Figure 2 illustrates examples of the applied preprocessing and augmentation techniques, including geometric transformations, color perturbations, and enhancement methods. The visual examples highlight how the preprocessing pipeline progressively improves the visibility of retinal structures while preserving anatomical integrity.

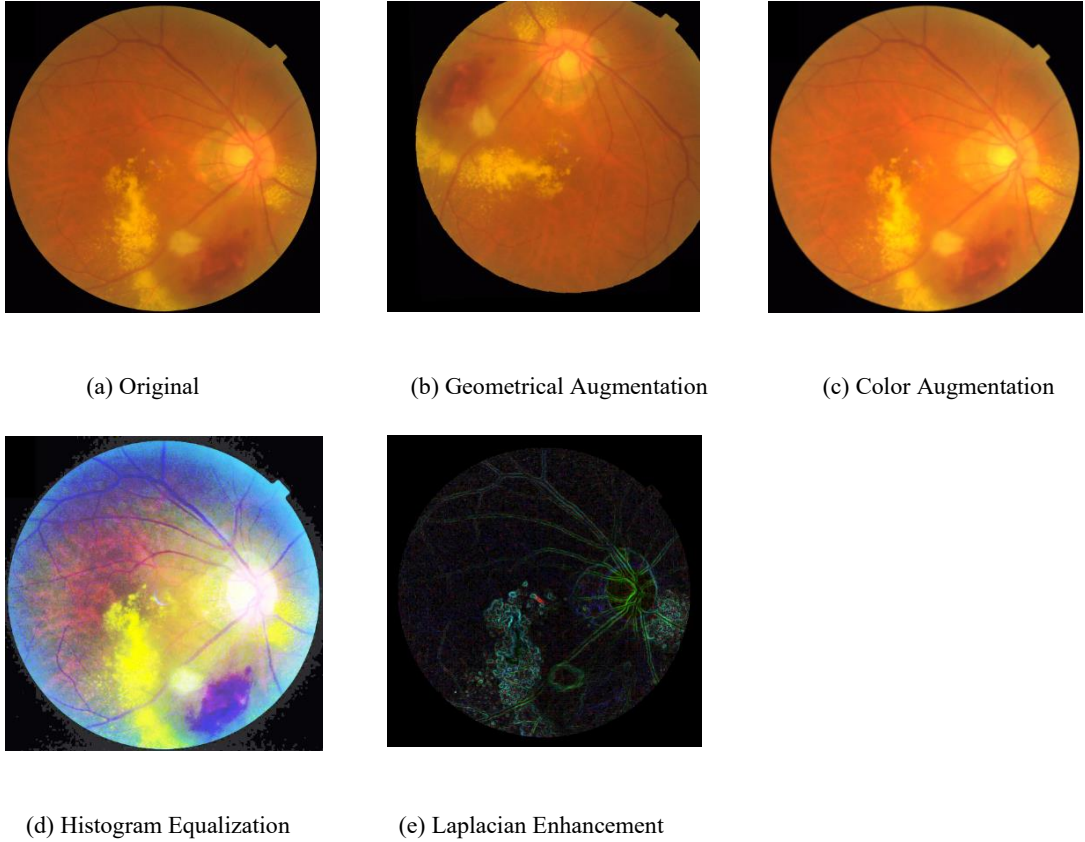

(a) Original  (b) Geometrical Augmentation  (c) Color Augmentation

(d) Histogram Equalization  (e) Laplacian Enhancement

**Figure 2:** Examples of the preprocessing and augmentation techniques applied to the fundus image datasets. The figure demonstrates successive stages of (a) geometric transformations (random rotation, translation, and flipping), (b) color perturbations (brightness and contrast jittering, Gaussian blur), and enhancement methods ( (c) histogram equalization and (d) Laplacian filtering). The image was taken from the FIVES dataset.

## 2.2.3. Training Procedures

To ensure consistency and comparability across datasets, we employed stratified sampling based on both dataset origin and class labels. This approach maintained balanced class distributions within training, validation, and test subsets, preventing potential biases caused by uneven sampling. The training results were stored separately for each dataset, allowing us to quantitatively assess domain shifts and evaluate cross-dataset generalization.

The ViT model was fine-tuned using the Adam optimizer with an initial learning rate of 0.00001 over 30 epochs. A cosine learning rate schedule was applied with five warm-up epochs, enabling gradual adaptation during the early training phase to stabilize convergence. Since the model was initialized with ImageNet-pretrained weights, the model did not require a lot of epochs for fine-tuning. To ensure full reproducibility, the same random seed was used across all devices, guaranteeing identical data splits and augmentation sequences across experiments. Each preprocessing setup (described in Section 2.2.2) was evaluated independently, ensuring that performance differences arose solely from the applied augmentation or enhancement techniques.

Following the identification of the best-performing preprocessing setup, we trained the constrained attention mechanism as defined in equation 1 using a U-Net based masking network. The pretrained ViT classifier and the U-Net were trained simultaneously which generated an attention mask that selectively preserves informative regions while suppressing irrelevant background. The model was optimized using the same training configuration as the ViT fine-tuning stage, Adam optimizer with an initial learning rate of 0.00001 and cosine scheduling. This training scheme allows the U-Net to refine its mask output based on the classifier's gradients, leading to a spatially interpretable representation of regions that contribute most significantly to classification decisions.

In contrast to the pretrained models, the GANomaly network was trained from scratch for 100 epochs, as no pretrained weights were available. The model was optimized using the Adam optimizer with an initial learning rate of 0.00001. The overall loss function consisted of a weighted combination of the reconstruction loss (weighted by 50), the adversarial feature map loss (weighted by 1), the latent space recreation loss (weighted by 1), and the normalized KL regularization (weighted by 0.3). The GANomaly was only trained on the healthy cohort, reducing the number of available samples significantly.

# 3. Results

## 3.1 ViT Performance

Overall, the ViT model demonstrated robust classification performance across all tested augmentation strategies, with total accuracies ranging from 0.789 to 0.843 on the test set. Table 2 summarizes key performance metrics for each dataset, providing insight into the model's generalization capabilities across diverse image sources. Among the evaluated configurations, the geometric augmentation setup achieved the highest overall accuracy (0.843), surpassing both the baseline without augmentation (0.825) and the contrast-enhanced variants. Histogram equalization reached 0.789, while Laplacian enhancement achieved 0.815, both underperforming relative to the non-enhanced baseline. These results suggest that although contrast and edge-enhancement methods can visually emphasize retinal vasculature and optic disc features, they may inadvertently introduce intensity distributions that are non-physiological, thereby slightly degrading model performance. Inclusion of brightness and color augmentations led to marginal changes in raw accuracy (0.841-0.843), but 10-fold cross-validation with a paired t-test confirmed that these differences were not statistically significant ($p > 0.05$), indicating comparable performance across these augmentation schemes. Table 3 presents the multi-class confusion matrix for the full test set. The results indicate that glaucoma and normal images were the most challenging to distinguish, reflecting the subtle structural changes in early-stage glaucoma and the similarity of certain normal anatomical variations to mild pathological features.

**Table 2**: Final test set performance across various augmentation strategies. Within each cell, results are reported in the following dataset order: AEyeDB, FIVES, Mendeley, Papila, and Messidor. Performance is compared using accuracy, weighted F1-score, and Matthews correlation coefficient (MCC) as evaluation metrics.

|  | Accuracy | Weighted F1 | MCC |
| --- | --- | --- | --- |
| No Augmentation | 1.000<br>0.842<br>0.800<br>0.825<br>0.944 | -<br>0.840<br>0.799<br>0.786<br>0.950 | -<br>0.787<br>0.732<br>0.350<br>0.836 |
| Geometric Augmentation | 1.000<br>0.741<br>0.839<br>0.857<br>0.911 | -<br>0.737<br>0.839<br>0.844<br>0.920 | -<br>0.650<br>0.784<br>0.512<br>0.759 |
| Color Augmentation | 1.000<br>0.730<br>0.833<br>0.841<br>0.956 | -<br>0.722<br>0.835<br>0.822<br>0.961 | -<br>0.634<br>0.779<br>0.444<br>0.865 |
| Histogram Equalization | 1.000<br>0.652<br>0.769<br>0.905<br>0.922 | -<br>0.653<br>0.770<br>0.893<br>0.924 | -<br>0.546<br>0.692<br>0.693<br>0.775 |
| Laplace Enhancement | 1.000<br>0.708<br>0.798<br>0.841<br>0.967 | -<br>0.708<br>0.801<br>0.812<br>0.966 | -<br>0.605<br>0.736<br>0.431<br>0.894 |

**Table 3**: The table shows the confusion matrix on the withhold test set for the model trained with geometric augmentation.

| | Normal | DR | Glaucoma | AMD | MS | RP | DE |
|---|---|---|---|---|---|---|---|
| Normal | 287 | 10 | 34 | 1 | 2 | 0 | 0 |
| DR | 5 | 245 | 2 | 1 | 6 | 2 | 4 |
| Glaucoma | 34 | 10 | 174 | 1 | 10 | 1 | 0 |
| AMD | 0 | 2 | 0 | 24 | 0 | 0 | 0 |
| MS | 3 | 7 | 5 | 1 | 47 | 3 | 1 |
| RP | 0 | 1 | 0 | 0 | 0 | 20 | 0 |
| DE | 2 | 2 | 0 | 0 | 0 | 1 | 14 |

## 3.2 Constrained Attention Mask

Table 4 presents the confusion matrix of the masked ViT model on the test set. Although the integration of the attention-guided U-Net slightly reduced overall performance with an accuracy of 0.832 compared to the baseline ViT model, a 10-fold cross-validation with a paired t-test indicated no statistically significant difference between the two configurations ($p > 0.05$). This suggests that the attention mask functions as intended without adversely affecting classification accuracy.

A detailed experimental comparison of pathological region localization obtained from the attention mask, Grad-CAM, integrated gradients, occlusion analysis, and anomaly detection is provided in Appendix B.

**Table 4**: The table shows the confusion matrix on the withhold test set for the model trained with geometric augmentation and constrained attention.

|  | Normal | DR | Glaucoma | AMD | MS | RP | DE |
| --- | --- | --- | --- | --- | --- | --- | --- |
| Normal | 305 | 3 | 23 | 1 | 2 | 0 | 0 |
| DR | 13 | 229 | 9 | 0 | 9 | 1 | 4 |
| Glaucoma | 45 | 8 | 163 | 1 | 12 | 1 | 0 |
| AMD | 0 | 3 | 1 | 22 | 0 | 0 | 0 |
| MS | 3 | 5 | 7 | 1 | 48 | 2 | 1 |
| RP | 1 | 0 | 0 | 0 | 0 | 20 | 0 |
| DE | 2 | 2 | 0 | 0 | 0 | 1 | 14 |

## 3.3 Anomaly Detection

The evaluation of the GANomaly based anomaly detection framework showed that neither the incorporation of KL regularization nor the addition of the attention-mask guided reconstruction loss resulted in measurable performance gains on the test set. Across all configurations, the anomaly detector consistently achieved an AUC of 0.76, indicating comparable ranking performance regardless of regularization strategy. However, their training dynamics differed: introducing stochasticity via the KL term in combination with the adversarial objective produced visibly noisier and less stable loss trajectories, as illustrated in Figure 3a, whereas deterministic configurations converged more smoothly.

Figure 3b presents the distribution of anomaly scores from the KL-regularized model. These scores were subsequently transformed into calibrated abnormality probabilities using GUESS, which applies Bayesian posterior estimation to map raw anomaly magnitudes to interpretable probability values. This calibration step enables the model to output continuous estimates reflecting the likelihood of pathology, thereby avoiding reliance on discrete threshold selection.

Given the objective of producing well-ordered probabilistic predictions we assessed performance primarily using the AUC, which measures ranking quality rather than threshold-specific accuracy. This metric is particularly appropriate for anomaly detection systems and supports clinical use cases where decision thresholds may vary depending on risk tolerance or screening context.

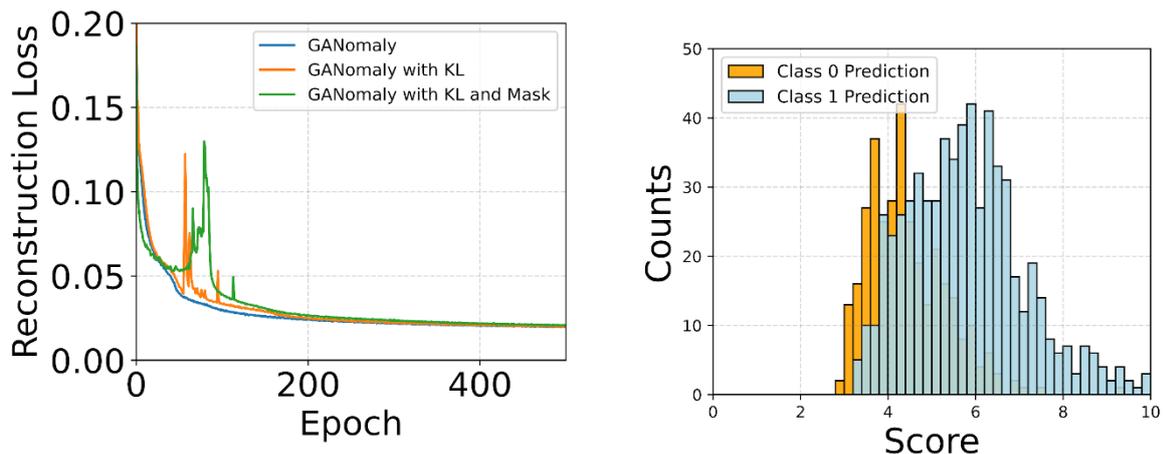

(a) Reconstruction Loss Training Curve  (b) Anomaly Score Distribution

**Figure 3:** The reconstruction loss training curves for the three setups are shown in figure (a) while figure (b) shows the distribution of the uncalibrated anomaly score on the test set.

A comparative evaluation on the unseen JSIEC dataset is summarized in table 5, contrasting the performance of the baseline ViT classifier with that of the GANomaly anomaly detector. For the ViT model, class-level predictions are reported directly, whereas for the GANomaly model, we present the mean calibrated abnormality probabilities obtained through GUESS for each disease category. This allows for a direct comparison between deterministic class assignments and probabilistically derived anomaly estimates.

Figure 4 visualizes the regions contributing most strongly to the anomaly signal by displaying the 90th-percentile absolute reconstruction errors for several pathological patterns. These maps highlight the correspondence between elevated reconstruction error and clinically relevant structures, offering qualitative insight into the detector's failure modes and emphasizing its potential utility as a weakly supervised localization tool.

**Table 5:** The table provides the corresponding fractions of classification for the ViT base model and the mean probabilities of abnormalities for the GANomaly model calibrated with GUESS.

|  | Base Model | Mean Probability of an Anomaly for GANomaly |
|---|---|---|
| Normal | Normal: 1.00 | 0.71±0.09 |
| DR | DR: 1.00 | 0.76±0.12 |
| Fibrosis | AMD: 0.20<br>DR: 0.70<br>MS: 0.10 | 0.81±0.11 |
| Pathological Myopia | AMD: 0.35<br>DR: 0.26<br>Glaucoma: 0.09<br>MS: 0.30 | 0.94±0.07 |
| Optic Atrophy | Normal: 0.17<br>DR: 0.17<br>Glaucoma: 0.66 | 0.76±0.05 |
| Laser Spots | Normal: 0.05<br>DE: 0.05<br>DR: 0.45<br>Glaucoma: 0.05<br>RP: 0.4 | 0.87±0.07 |

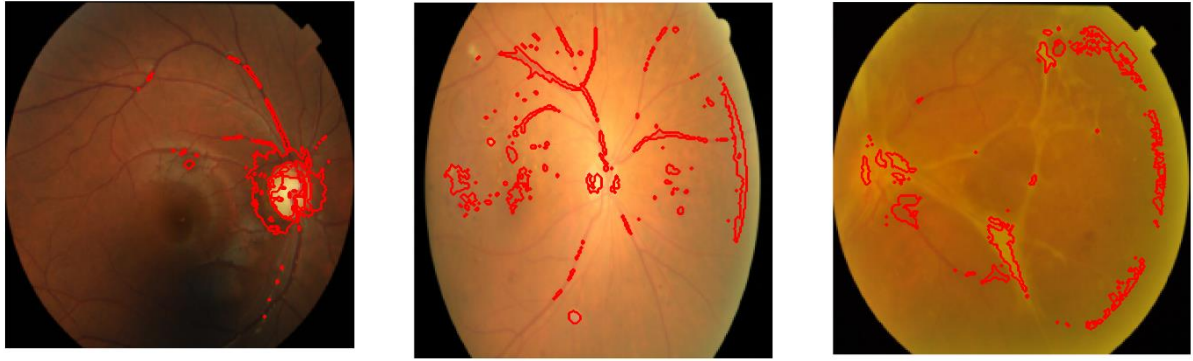

(a) Healthy Image            (b) DR Image            (c) Fibrosis Image

**Figure 4:** Visualization of the 90th-percentile absolute reconstruction error maps for representative pathological cases. These maps highlight the image regions contributing most strongly to the anomaly signal, illustrating the correspondence between elevated reconstruction error and clinically relevant structures.

## 4. Discussion

The performance of the ViT across multiple augmentation strategies highlights both the potential and the current challenges of fundus-based disease classification. Although the overall accuracy remains high across all configurations, the observed variability in weighted F1 and MCC underscores the difficulty of achieving balanced performance across all disease categories, an expected outcome in inherently imbalanced medical datasets .

Early-stage disease detection remains particularly challenging, as reflected in the confusion matrix. While the ViT model reliably distinguishes healthy from pathological images, it exhibits class-specific variability. DR and AMD classes demonstrate strong sensitivity and consistent classification performance, whereas glaucoma shows the highest rate of misclassification, often being predicted as normal. This observation suggests that early glaucomatous changes, which primarily manifest as subtle alterations in the optic nerve head and peripapillary regions, remain difficult to capture even for high-capacity transformer-based models.

Interestingly, no single augmentation strategy yielded a statistically significant improvement across all datasets. Both geometric and color augmentation produced comparable results, suggesting that the model effectively leverages both spatial and chromatic cues. However, histogram equalization achieved the highest accuracy within the Papila dataset, composed exclusively of normal and glaucoma images, indicating that contrast normalization may enhance feature discrimination when subtle textural and structural variations dominate. In contrast, Laplacian enhancement and heavy color perturbations tended to reduce performance, likely due to oversharpening and the artificial amplification of irrelevant image gradients.

When evaluating the ViT model with geometric augmentation on the Papila dataset, the model achieved an AUC of 0.91. This represents a substantial improvement compared to the results reported in the original Papila study [33], which achieved an AUC of approximately 0.87 using a conventional convolutional network ensemble. The increase can be attributed to the use of a state-of-the-art transformer-based architecture, the integration of multiple datasets during training which enriches the learned feature distribution, and the application of effective data augmentation strategies that improve generalization across unseen samples.

Differences in dataset-level performance further highlight the impact of domain shift, as illustrated by the training curves presented in Appendix A. Datasets characterized by consistent acquisition protocols and manual quality control, such as FIVES and AEyeDB, achieved the highest performance, indicating that standardized imaging conditions and reduced intra-dataset variability substantially enhance model generalization. In contrast, the particularly strong results on AEyeDB may also reflect overfitting to dataset-specific image characteristics, emphasizing the need for future work on cross-dataset normalization and domain adaptation techniques.

Although the constrained attention mechanism did not improve classification accuracy, its integration represents a meaningful step toward model-intrinsic interpretability. Unlike post-hoc methods such as Grad-CAM or occlusion, which rely on surrogate explanations that may be biased by baseline image choices [39], our U-Net derived spatial masks are optimized jointly with the classifier. Despite the minor performance decrease, this approach offers a more principled way of enforcing spatial consistency and may benefit clinical interpretability in future extensions.

The GANomaly-based anomaly detector achieved a consistent AUC of 0.76, independent of whether KL regularization or mask-based reconstruction loss was applied. Unlike supervised classifiers, anomaly detectors are mathematically grounded to generalize to unseen pathology distributions. The calibrated probabilities obtained through GUESS provide interpretable outputs and avoid threshold biases. Moreover, the reconstruction-error maps highlight clinically meaningful structures, offering an inherently explainable signal that points to suspicious regions without requiring class labels.

Unlike supervised classifiers, anomaly detectors are mathematically grounded to generalize to unseen pathology distributions. The calibrated probabilities obtained through GUESS provide interpretable outputs and avoid threshold biases. Moreover, the reconstruction-error maps highlight clinically meaningful structures, offering an inherently explainable signal that points to suspicious regions without requiring class labels.

Future work should prioritize increasing the number of healthy training images, as the current imbalance biases the probabilistic calibration toward higher abnormality likelihoods. Additionally, domain adaptation and dataset harmonization will be essential to mitigate dataset-specific biases.

## 5. Conclusion

This work demonstrates the strong potential of transformer-based architectures for retinal disease classification while emphasizing enduring limitations related to subtle pathology detection and cross-domain robustness. The ViT achieved high performance across multiple datasets and augmentation strategies, with particularly reliable detection of DR and AMD. Glaucoma remained the most difficult class, reflecting the subtle morphological changes associated with early disease. No augmentation strategy yielded a universal improvement. Geometric and color augmentations were consistently beneficial, whereas Laplacian enhancement reduced discriminability probably due to oversharpening artifacts. Histogram equalization was advantageous only in datasets dominated by nuanced structural cues, such as Papila.Our cross-dataset evaluation revealed the substantial influence of acquisition quality and dataset consistency on model generalization.

The GANomaly-based anomaly detector, although less accurate than the ViT due to limited training diversity and lower model capacity, offered two distinct advantages: intrinsic explainability through reconstruction-based error maps and stable generalization to unseen datasets. Its GUESS-calibrated probabilistic outputs avoid threshold tuning and provide a mathematically grounded alternative to classifier confidence scores. This makes anomaly detection a valuable complementary strategy, especially in settings where disease categories are open-ended or incompletely represented during training.

Future work will focus on expanding the pool of healthy fundus images to improve calibration stability, reducing domain shift through harmonization and self-supervised pretraining, and exploring hybrid architectures that combine discriminative transformer models with generative anomaly detection. Together, these directions aim to advance trustworthy, explainable, and generalizable AI tools for ophthalmic screening.

**Ethics Statement**
All research complied with German legal regulations. The study was approved by the Ethics Committee of the Heinrich Heine University on November 8, 2023 (ID: 2023-2622). Informed consent was obtained from all participants.

**Acknowledgements**
This work is partially funded by hessian.AI in the Connectom project VirtualDoc. We also wish to thank the students of the Heinrich Heine University Düsseldorf who participated in the study. The MESSIDOR dataset was kindly provided by the Messidor program partners.

**Author Contributions**
D. H. served as the principal investigator, overseeing the study's design and implementation. B. F. and T. P. provided valuable insights during the analysis phase. J. B. R. led the study's execution, analysed the data, and conceptualized the study. A. C., J. S., and N. E. contributed to the ethical proposal, study implementation, and medical interpretation.

**Data Availability Statement**
As outlined in the study description (available: https://drks.de/search/de/trial/DRKS00033094), we are committed to adhering to the data security policies of both the European Union (EU) and Germany. Accordingly, any researcher interested in accessing the data may do so by submitting valid research comply with the data security regulations of both the EU and Germany. This process ensures the confidentiality and integrity of the data while facilitating responsible and ethical research practices

# Appendix A. Network Design and Ablation Details

## A.1. Network Design

**Table B.6:** Architectural summary of the proposed U-Net model used for constrained attention mask generation. The network integrates residual blocks with attention at the middle section, featuring symmetric encoder-decoder paths with skip connections. Downsampling and upsampling operations control spatial resolution, while a final sigmoid activation constrains the output mask to the range [0,1].

| Stage | Layer / Block | Input Channels | Output Channels | Key Features / Notes |
|---|---|---|---|---|
| Input | AdaptiveNorm + Conv2d | 3 | 32 | Input normalization, 3×3 conv, padding=1 |
| Down 1 | Downsample | 32 | 32 | Residual connections |
| Down 2 | Downsample | 32 | 64 | Residual block |
| Down 3 | Downsample | 64 | 128 | Residual block |
| Down 4 | Identity | 128 | 128 | No downsampling |
| Middle | ResBlockWithAtt | 128 | 128 | Attention always applied, residual connections |
| Up 4 | Upsample | 256 (skip+up) | 128 | Concatenates skip connection from Down 4 |
| Up 3 | Upsample | 256 | 64 | Concatenates skip connection from Down 3 |
| Up 2 | Upsample | 128 | 32 | Concatenates skip connection from Down 2 |
| Up 1 (last) | Identity | 64 | 32 | Concatenates skip connection from Down 1 |
| Output | Conv2d + Sigmoid | 32 | 1 | Produces single-channel mask, pixel values in [0,1] |

## A.2. Ablation

Figure B.5 illustrates the learning curves for the different augmentation strategies. In figure B.6 the accuracies on the validation set for each individual dataset is listed.

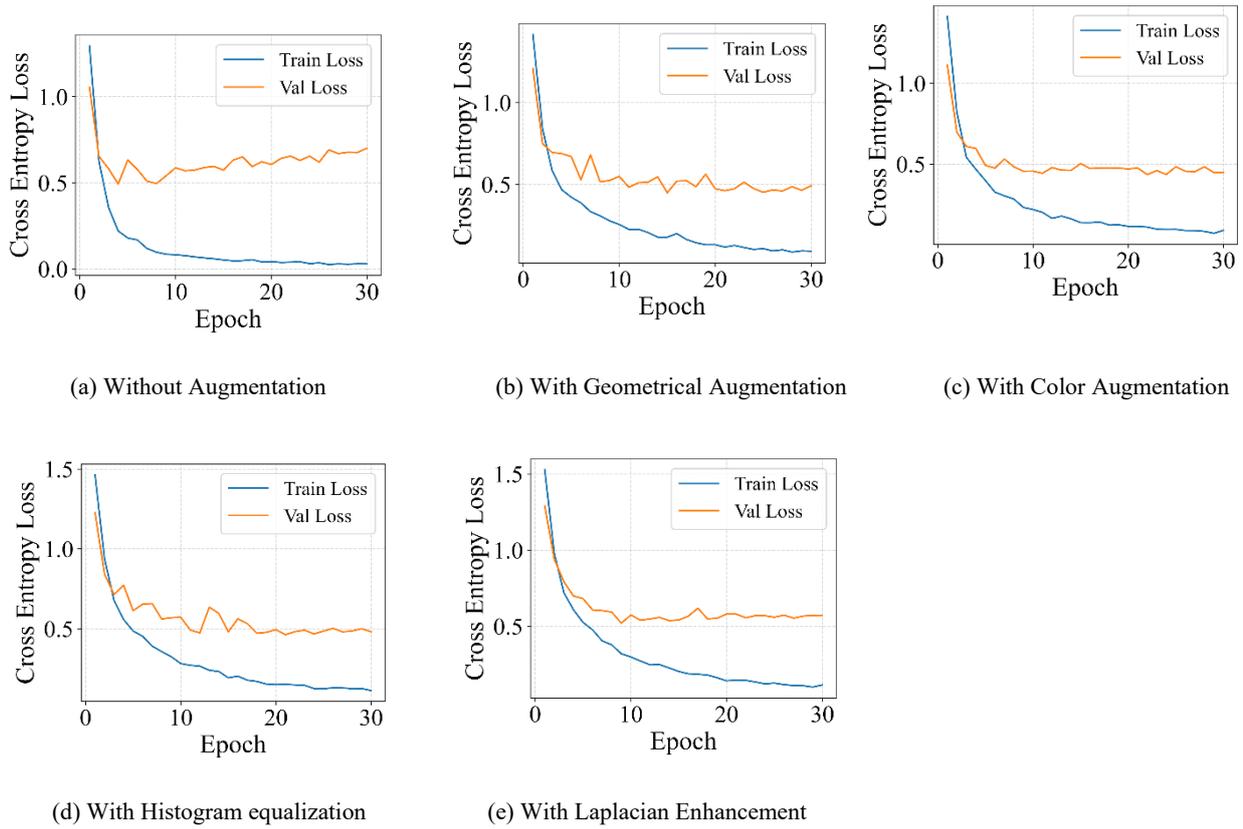

(a) Without Augmentation  (b) With Geometrical Augmentation  (c) With Color Augmentation

(d) With Histogram equalization  (e) With Laplacian Enhancement

**Figure B.5:** Learning Curves for different augmentation strategies.

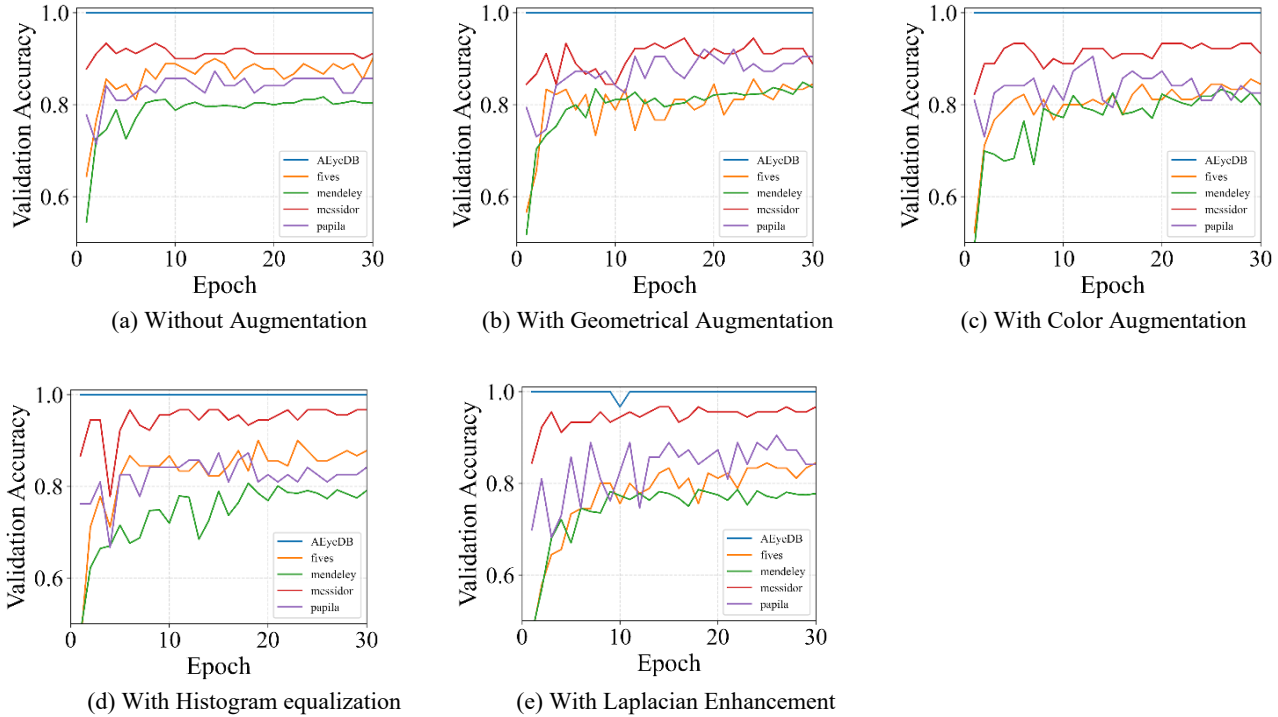

**Figure B.6:** The figure illustrates the validation accuracy per dataset to observe any domain shifts.

# Appendix B. Model Explainability

Figure B.7 presents a comparison of several explainable AI techniques, highlighting their respective limitations when applied to fundus image interpretation. The first method evaluated was patch-based occlusion, in which image regions are systematically masked and the resulting changes in model prediction are measured. We report the 90th-percentile impact scores across the two patch sizes 16×16 (matching the ViT patch resolution) and 8×8 (providing finer spatial granularity). However, the resulting saliency patterns exhibit highly erratic and spatially inconsistent behavior, limiting the clinical interpretability and reliability of this approach.

We further assessed integrated gradients and Grad-CAM, which derive localization from gradient-based changes. While CAM is well suited for CNNs with designated feature aggregation layers, its application to ViTs is less straightforward because no single layer naturally encodes spatial importance, leading to unstable or layer-dependent results. SHAP-like methods present an additional challenge: they require a predefined baseline, and simple choices such as black or white images may introduce unintended biases or unrealistic reference states.

For comparison, we also include the constrained attention mask proposed in this work. Unlike post-hoc explainable AI methods, it provides a stable and model-intrinsic representation of spatial relevance that is learned jointly during training.

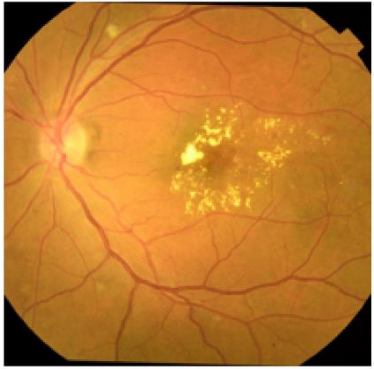
(a) Original

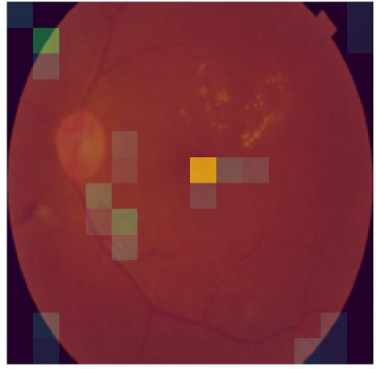
(b) 8×8 Occlusion

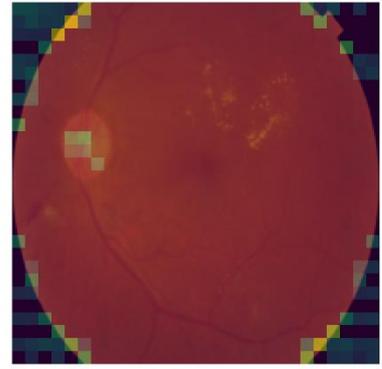
(c) 16×16 Occlusion

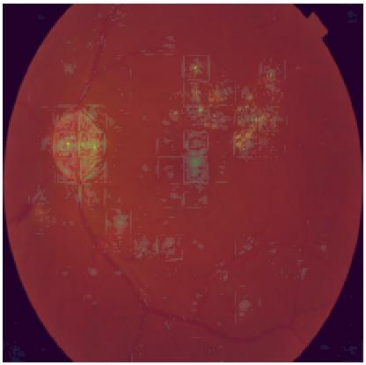
(d) Integrated Gradients

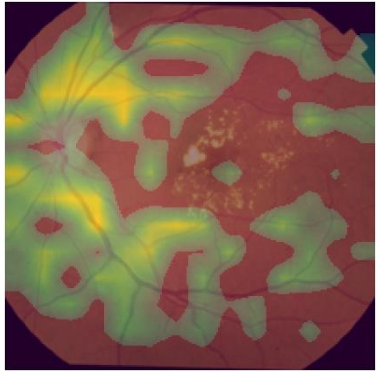
(e) Grad-CAM

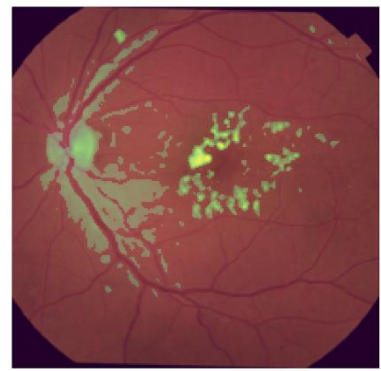
(f) Attention Mask

**Figure B.7:** Comparison of explainable AI methods including occlusion analysis (8×8 and 16×16 patches), gradient-based CAM, integrated gradient contribution, and the proposed constrained attention mask. The visualization highlights the instability and methodological limitations of traditional post-hoc approaches, in contrast to the consistency of the integrated attention-based method. The image was taken from the Mendeley dataset and contains DR pathologies.